# TablEye: Seeing small Tables through the Lens of Images

Seung-Eon Lee
College of Transdisciplinary Studies
Daegu Gyeongbuk Institute of Science & Technology(DGIST)
Daegu 42988, Korea
tmddjs7531@dgist.ac.kr

Sang-Chul Lee
Division of Nanotechnology
Daegu Gyeongbuk Institute of Science & Technology(DGIST)
Daegu 42988, Korea
sangchul.lee@dgist.ac.kr

## ABSTRACT

The exploration of few-shot tabular learning becomes imperative. Tabular data is a versatile representation that captures diverse information, yet it is not exempt from limitations, property of data and model size. Labeling extensive tabular data can be challenging, and it may not be feasible to capture every important feature. *Few-shot tabular learning*, however, remains relatively unexplored, primarily due to scarcity of shared information among independent datasets and the inherent ambiguity in defining boundaries within tabular data. To the best of our knowledge, no meaningful and unrestricted few-shot tabular learning techniques have been developed without imposing constraints on the dataset. In this paper, we propose an innovative framework called **TablEye**, which aims to overcome the limit of forming prior knowledge for tabular data by adopting domain transformation. It facilitates domain transformation by generating tabular images, which effectively conserve the intrinsic semantics of the original tabular data. This approach harnesses rigorously tested few-shot learning algorithms and embedding functions to acquire and apply prior knowledge. Leveraging shared data domains allows us to utilize this prior knowledge, originally learned from the image domain. Specifically, TablEye demonstrated a superior performance by outstripping the TabLLM in a *4*-shot task with a maximum *0.11* AUC and a STUNT in a *1*-shot setting, where it led on average by *3.17*% accuracy.

## 1 Introduction

It is a common misperception that a large volume of data is indispensable for the deep learning techniques with a neural network. Indeed, dataset size plays a critical role in enhancing model performance. Regardless of a neural network model quality, it seems futile without access to ample data. This data insufficient problem frequently arises due to some reasons such as high costs, privacy concerns, or security issues. Despite these challenges, there are various attempts to improve accuracy through deep learning with limited labeled data. This line of research is known as *few-shot learning* [1].

Few-shot learning models have made significant strides in practical applications within the image domain [2] and language domain [3]. These advancements can be attributed to the relative ease of leveraging large-scale datasets that have already been collected, such as ImageNet [4] and internet-based corpora, to establish prior knowledge.

Few-shot learning in the tabular domain [5], however, has received relatively little attention. The lack of research in this area can be traced back to several factors. Firstly, compared to the image and language domains, tabular datasets lack shared information [5]. Unlike images or language data, where prior knowledge can be learned from related examples within the different datasets, it is challenging to establish similar relationships in tabular data. For example, while learning to distinguish between dogs and lions may assist in distinguishing between cats and tigers, learning to predict solar power generation will not necessarily aid in understanding trends in the financial market. Secondly, defining clear boundaries for tabular data is a complex task. Image and language data possess physical or visual representations, allowing their boundaries to be defined by parameters such as pixels, color channels (R, G, B), image size, vocabulary(words), and grammar. In contrast, tabular data lacks a distinct shared representation. Various features within tabular data have independent distributions and ranges, and missing values may be present. As a result, even completely identical values in different datasets can carry entirely different meanings.

Attempts at previous few-shot learning in the tabular domain have embraced two approaches. The first approach involves utilizing large unlabeled tabular datasets to generate tasks and learn prior knowledge through semi-few-shot learning [6]. This method requires, however, additional well-composed unlabeled dataset, and its effectiveness is influenced by the size of the unlabeled dataset [7]. The second approach involves leveraging template-generated serialized tabular data, using prior knowledge from the language domain, and feeding it to a Large Language Model (LLM) in a prompt-based format [8]. This approach necessitates meaningful feature names for serialization, imposes restrictions on the number of features, and demands high computing power for fine-tuning and operating the LLM.

This paper is motivated by these challenges and presents a novel framework, **TablEye**, which facilitates the seamless implementation of few-shot tabular learning. It offers unrestricted usage while ensuring high accuracy. TablEye assumes that prior knowledge learned in the image domain can help solve any task in tabular domain. For an intuitive example, if a child learns about an apple from a picture, they can connect it to a letter ("*Apple*") and a number ("*An*" or "*One*") and make associations. If additional information, such as rules or relationships between numbers, is provided, the child can infer that two apples are present by observing two apple photos side by side. However, for a child who has only learned the numbers "*1*" and "*2*", understanding that *1 + 1* equals *2* may not come easily. Similarly, if we incorporate information about tabular data into neural networks trained solely on images, even a small labeled data can yield superior performance compared to traditional machine learning approaches that rely on larger amounts of labeled data.

To experimentally validate our assumptions, we adopt a few-shot learning algorithms that have already demonstrated good performance in the image domain. To utilize prior knowledge formed in the image domain, we transform each row of the tabular data into an image format that the model can comprehend. We extend the application of spatial relations to three channels and convert the tabular data into a format similar to other images, preserving the meaning of the tabular data while adopting the structure of an image.

Our proposed approach achieves comparable or even superior performance to previous research through experiments on various datasets. Moreover, it offers the flexibility to perform few-shot learning tasks without being constrained by composition of dataset. TablEye overcomes the need for large unlabeled datasets by leveraging the image domain, and it requires less computing cost due to its smaller model size than one of the LLM. To the best of our knowledge, this paper represents the first attempt to apply prior knowledge from the image domain to few-shot learning in the tabular domain. The proposed few-shot tabular learning technique has the potential to provide artificial intelligence models that can achieve accurate results with only a small amount of data in scenarios where data acquisition is challenging, such as disease diagnosis in the medical industry.

The main **contributions** of this work are:

- This work represents the first attempt to leverage large image data as prior knowledge to address the problem of few-shot tabular learning, formation of prior knowledge.
- We propose a novel framework, **TablEye**, which employs domain transformation to apply prior knowledge from image data to few-shot tabular learning.
- We have successfully overcome the limitations associated with existing few-shot tabular learning models, including constraints related to feature size of dataset, the requirement for large quantities of unlabeled data, and the demand for extensive computational resources.

## 2 Background

### 2.1 Weakly Supervised Learning

Weakly supervised learning [9] is a type of supervised learning that utilizes labeled data that is incomplete, inaccurate, or contains noise. Definitionally, weakly supervised learning only encompasses classification and regression, whereas few-shot learning encompasses reinforcement learning and incorporates prior knowledge. In certain cases, the weakly labeled data can be transformed into labeled data, providing a solution to the few-shot learning problem.

### 2.2 Transfer Learning

The transfer learning [10] paradigm in machine learning involves utilizing related tasks with ample labeled data as source tasks to improve the performance of a target task with limited labeled data. Transfer learning employs prior knowledge acquired from other data sets, making it similar to few-shot learning and multi-task learning [11]. Transferring prior knowledge from source tasks to target tasks in few-shot learning is a popular method in transfer learning. Examples include cross-recommendation[12] and Wi-Fi localization[13].

### 2.3 Few-Shot learning

The Few-shot Learning problem is a type of machine learning problem characterized by having a limited amount of labeled training data. Equation (1), (2), (3) show the relation. It is particularly relevant in situations where the data set containing the train set $D_{train}$ and the test set $D_{test}$, $D_{train}$ containing the sample pairs $(x_i, y_i)$ with $i$ ranging from *1* to small size *I*, and $D_{test}$ consisting of $x^{test}$. Applications of few-shot learning include image classification, sentiment classification, and object recognition. The general notation of *N*-way-*K*-shot classification refers to a train set composed of *N* classes, each with *K* examples, totaling I=NK examples.

$$D = \{ D_{train}, D_{test}\} \qquad (1)$$

$$D_{train} = \{(x_i, y_i)\}_{i=1}^{I} \qquad (2)$$

$$D_{test} = \{x^{test}\} \qquad (3)$$

The objective of solving the few-shot learning problem is to find an optimal hypothesis $\hat{h}$, such that $\hat{y} = \hat{h}(x; \theta)$, in the hypothesis space $H$ ($\hat{h} \in H$). This can be a challenging task due to the large hypothesis space $H$ and the small size of the labeled set *I*, which makes traditional supervised learning techniques, which require a large amount of labeled data, unfeasible. Equation (4), (5) show the relationship of expected risk $R$ and empirical risk $R_I$ with *x, y*. To minimize the expected risk $R$, we approximate it using the empirical risk $R_I$ [14][15]., since the true underlying probability distribution $p(x, y)$ is unknown.

$$R(h) = \int L(h(x), y) dp(x, y) \qquad (4)$$

$$R_I(h) = \frac{1}{I}\sum_{i}^{I} L(h(x_i), y_i) \qquad (5)$$

However, finding the optimal hypothesis $\hat{h}$ remains elusive. To address this challenge, we decompose the total error of the problem by using the best approximation $h^*$ for $\hat{h}$ as follows[16][17]. Equation (6) shows the decomposition of total error. *H* is the size of hypothesis space, and *I* is the size of training set.

$$E[R(h_I) - R(h^*)] = E[R(h^*) - R(\hat{h})] + E[R(h_I) - R(h^*)]$$
$$= \varepsilon_{approximation}(H) + \varepsilon_{estimation}(H, I) \quad (6)$$

According to the formula (6), the total error is comprised of the approximation error, $\varepsilon_{approximation}$ of the hypothesis space $H$ and the estimation error, $\varepsilon_{estimation}$ of our hypothesis $h^I$ with the best approximation $h^*$. The formula highlights that the size of the hypothesis space and the size of the training set are crucial factors in few-shot learning. As a result, there are generally three approaches to solving few-shot learning. First, increasing the size of the training set, $I$. Second, defining the hypothesis space $H$ from the perspective of the model[18, 19, 20]. Third, changing search strategy for the best approximation $h^*$ within the hypothesis space from the perspective of the algorithm. The approach from the perspective of data involves augmenting the small training set by leveraging prior knowledge, such as through transforming the data samples from the training set, weakly labeled set, unlabeled set, or similar datasets. The approach from the perspective of the model reduces the size of the hypothesis space by utilizing prior knowledge through multi-task learning or embedding learning. Lastly, the approach from the perspective of the algorithm adjusts the strategy for searching for the optimal hypothesis $\hat{h}$ within the hypothesis space $H$, such as by refining parameters or learning the optimizer.

## 3 Related Works

### 3.1 Semi-Few-Shot Tabular Learning: STUNT

STUNT[6] represents a semi-few-shot learning technique aimed at enhancing the performance of tabular learning in scenarios with sparse labeled datasets, utilizing a substantial quantity of reasonably coherent unlabeled data. This method marks an attempt to resolve the few-shot learning problem from a data perspective, by learning prior knowledge from an unlabeled set to which arbitrary labels have been assigned. To generate these arbitrary labels, it adopted the K-means clustering technique. After the labeling of the unlabeled set, random columns are selected to create train tasks. This approach utilizes a Prototypical Network[21] to learn prior knowledge from these self-generated tasks, and it has demonstrated impressive performance. Both STUNT and TablEye, despite not yet having undergone substantial development, share a common focus on the industrially significant problem of few-shot tabular learning. This method, as a semi-few-shot learning technique, operates exclusively within the tabular domain, by the way requires a substantial quantity of reasonably consistent unlabeled data. The size of the unlabeled set also can significantly influence the performance of STUNT[7]. In contrast, our method learns prior knowledge from the image domain, a source from which data can be more readily procured. Furthermore, TablEye operates as a few-shot learning technique, functioning with only a small amount of labeled data. This distinction highlights the different approaches being taken within the field to address the challenges of few-shot tabular learning.

### 3.2 Few-Shot Tabular Learning: TabLLM

In the domain of few-shot tabular learning, the study TabLLM[8] offers a unique perspective by harnessing the pre-existing knowledge embedded within a Large Language Model (LLM). Characterized as a model-perspective methodology, it illuminates a different angle to the problem of few-shot learning[22]. The process employed by this method involves the conversion of original tabular data into a text format following a specific template, a process known as serialization. Table 1 offers the example of serialization. The text template includes the serialized tabular data. This transformation reformats tabular data into a more adaptable textual form, making it suitable as the prompt for LLM. Following the serialization, this data is utilized to fine-tune the LLM[23]. The prompts, derived from the serialized data, are processed by the model to produce the corresponding outputs. The T0 encoder-decoder model, equipped with an extensive set of *11 billion* parameters, plays a crucial role in this process [24]. This large parameter set, indicative of the extensive model training, also necessitates substantial computational resources, presenting a potential challenge. A key distinction between TablEye and this table-to-text method lies in the source of the prior knowledge applied. While we draw upon knowledge from the image domain, it relies on linguistic domain knowledge. This contrast underscores the variety of approaches being investigated in the field to address the challenge of few-shot learning. An important aspect of the LLM is its potential for learning categoric features. This potential is, however, conditional on the presence of meaningful feature names, which may limit its applicability in certain contexts. Further, the considerable size of model could pose computational difficulties. Notably, TabLLM is also constrained by its token length, which imposes a constriction on the number of features that can be processed in each experiment. Specifically, datasets with more than about *30* features exceed this model capacity, presenting another limitation to its application. In summary, TabLLM presents a novel approach to few-shot tabular learning through the application of linguistic domain knowledge despite some constraints related to meaningful feature name requirement, large model size, and token length limitations,

**Table 1. Example of TabLLM serialization using template**

| Original tabular data | | | | | |
|---|---|---|---|---|---|
| Buying | Maint | Doors | Persons | Lug_boot* | safety |
| Low | Low | Three | Four | medium | medium |

*Lug_boot is the trunk size.

| Text template |
|---|
| The Buying price is low. The Doors is three. The Maintenance costs is low. The Persons is more than four. The Safety score is medium. The Trunk size is medium. |

## 4 Our Approach

### 4.1 Overview

This research paper introduces a novel framework called **TablEye**, aimed at enhancing the effectiveness of few-shot tabular learning. Figure 1 shows the overview of TablEye. TablEye applies efficient few-shot learning algorithms in the image domain by performing domain transformation from the tabular domain to the image domain. The framework comprises two main stages: the *transformation stage* from the tabular domain to the image domain and the *prior knowledge learning stage* in the image domain. In the tabular domain, TablEye preprocesses tabular data and undergoes a transformation process into a three-channel image, referred to as a *"tabular image."* Subsequently, few-shot tabular classification is performed using prior knowledge learned from mini-ImageNet in the image domain. To generate tabular images from tabular data, a method based on feature similarity is employed, incorporating spatial relations into the tabular images. In the stage of learning prior knowledge, ProtoNet (Prototypical Network)[21] and MAML (Model-Agnostic Meta Learning)[25] are employed, as they demonstrate high performance and can be applied to various few-shot learning structures. The backbone for embedding and the classifier for the few-shot task are connected sequentially. During the process of learning embeddings in a dimension suitable for classification through the backbone, Cross-entropy loss is utilized.[26]

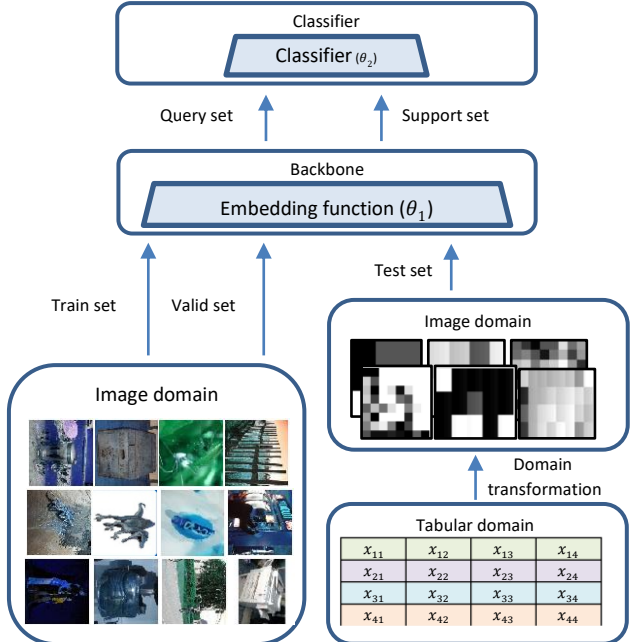

**Figure 1. Overview of TablEye. The natural images of image domain are part of mini-ImageNet[27].**

### 4.2 Domain Transformation

The domain transformation stage aims to convert tabular data into the desired form of images (*3, 84, 84*), while preserving the characteristics and semantics of the tabular data. We hypothesize that the difference between images and tabular data lies in the association with neighboring values and spatial relations[28]. The pixels in an image exhibit strong correlations with adjacent pixels, and this is why the kernels in a Convolutional Neural Network (CNN) play an important role. Therefore, we incorporate spatial relations into tabular data and undergo a process of shaping it into the desired form.

Given $N$ ($=N_r \times N_c$) features, we measure the Euclidean distance between these features and rank them to create an ($N, N$) feature matrix, denoted as $R$. We also measure the distance and rank between $N$ elements to generate an ($N, N$) pixel matrix, denoted as $Q$. The process of generating $R$ and $Q$ is described in detail in Appendix A. Then, we compute the Euclidean distance between $R$ and $Q$ and rearrange the positions of the features to minimize the distance, aiming to align the feature distance and pixel distance, thus assigning spatial relations. This results in obtaining a 2-dimensional image of size $N_r \times N_c$, where features with closer distances correspond to pixels that are closer to each other. In the equations below, $r_{ij}$ and $q_{ij}$ represent the elements at the $i$-th row and $j$-th column of $R$ and $Q$, respectively. By minimizing the distance between $R$ and $Q$ according to the equations, we align the feature distance and pixel distance, thus assigning spatial relations.

$$-Loss(R, Q) = \sum_{i=1}^{N} \sum_{j=1}^{N} (r_{ij} - q_{ij})^2$$

By repeating the same elements in a matrix $M$ of size $N_r \times N_c$, we obtain an image of size (*84, 84*). Applying the same (*84, 84*) image to each channel, we obtain an image of size (*3, 84, 84*). We refer to these images transformed from tabular data as the "tabular images." Figure 2 represents the results of transforming one data sample from each of the six datasets[29] used in the experiment into tabular images according to the proposed domain transformation method presented in this paper.

$$repeat(M, (84 \;//\; N_r + 1), axis = 0)$$
$$repeat(M, (84 \;//\; N_c + 1), axis = 1)$$
$$M = M[:84, :84]$$

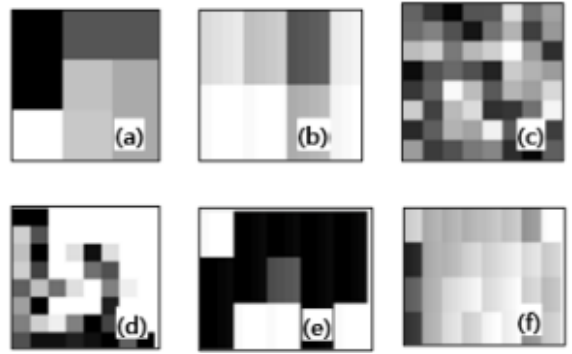

**Figure 2. Example tabular images. (a), (b), (c), (d), (e) and (f) are tabular images from CMC, Diabetes, Karhunen, Optdigits, Lung and Cancer data respectively.**

### 4.3 Learning Prior Knowledge

The proposed TablEye model consists of a backbone that serves as an embedding function to operate in the suitable dimension for few-shot learning, and a classifier that performs the few-shot learning task based on the embedded support set. TablEye utilizes mini-ImageNet[27] to train backbone and classifier We adopted four different backbone architectures as shown in Figure 3. It is because the structure and training state of the backbone can significantly impact the training state of the classifier. Figure 3

TablEye: Seeing small Tables through the Lens of Images

illustrates the actual architectures of the four backbones, namely Resnet12, Conv2, Conv3, and Conv4, proposed and experimentally validated in this paper. Hereinafter, Resnet12, Conv2, Conv3, and Conv4 refer to each backbone depicted in Figure 3 within this paper. Resnet12 is a complex and deep backbone with a *12*-layer ResNet[30] structure. Conv2, Conv3, and Conv4 are intuitive and shallow backbone architectures with *2, 3*, and *4*-layer CNN networks, respectively. The backbone continuously learns to achieve a better embedding function for the classifier based on the predictions of the classifier using cross-entropy loss.

The classifier plays a direct role in the few-shot learning task based on the embedded tabular images as latent vectors. In this paper, we adopt the principles of Prototypical Network[21], prototypes and inner loop adaptation of MAML[25] as our classifier. Both principles can be applied to various machine learning model structures. Moreover, recent studies have shown that few-shot learning with Prototypical Network achieves better performance than other complex few-shot learning models. Considering our goal of creating a model that operates with limited computing resources, we choose these two options for the classifier. When selecting the **Proto-layer** as the classifier, the classifier forms prototypes by averaging the latent vectors of the support sets for each class. It predicts the result by measuring the distances between the latent vectors of the query set and each prototype to determine the closest class. Alternatively, when selecting the **MAML-layer** as the classifier, we iteratively train a trainable fully connected layer within the inner loop using the latent vectors of the support set. The fully connected layer is then applied to the latent vectors of the query set to make predictions.

## 5 Experiment

### 5.1 Experimental Environment

To validate the hypothesis of this paper, we conducted experiments using image data from mini-ImageNet[27] and open tabular data from OpenML[29] and Kaggle. We constructed a train set consisting of *50,400* images and a validation set of *9,600* images from mini-ImageNet. For the test set, we composed tabular images after domain transformation. To ensure clear validation of the hypothesis, we applied the following criteria in selecting the tabular datasets for experiments: (1) Diversity of features: dataset containing only categorical features, dataset containing only numerical features, and dataset containing both categorical and numerical features, (2) Diversity of tasks: binary classification and multiclass classification, (3) Inclusion of medical data for industrial value validation. Table 2 presents the detail of the datasets used in the paper. The # indicates the number. The column named 'type' indicates the type of feature including num eric only feature, categoric only feature and numeric and categoric feature. The *N* and *C* of 'type' indicate numeric and categoric feature. The diabetes, Lung, Cancer and Prostate datasets are medical datasets. Throughout the research process, the main question was whether the prior knowledge learned from natural images in mini-ImageNet could be applied to tabular images. To address this, we employed t-SNE(t-Distributed Stochastic Neighbor Embedding)[31] technique to embed and visualize the distributions of natural images and transformed tabular images in a *2*-dimensional space. Figure 4 visually presents the results of embedding into a two-dimensional space using t-SNE[31]. Based on the *2*-dimensional embedding results, we measured the maximum distance, denoted as $distance_{max}$, from the mean vector of natural images as the center of two circles, $c_1$ and $c_2$. We then drew two circles: circle $c_1$ with a radius of $distance_{max}$ and circle $c_2$ with a radius of *0.8 \* $distance_{max}$* . The scattered points in Figure 4 represent individual datasets, while the red and blue circles represent c1 and c2, respectively. By comparing the tabular images with the two circles, we observed that some tabular images fell within $c_2$, while the majority of tabular images fell within $c_1$ . Therefore, we concluded that there is no domain shift issue in learning the prior knowledge of tabular images from natural images.

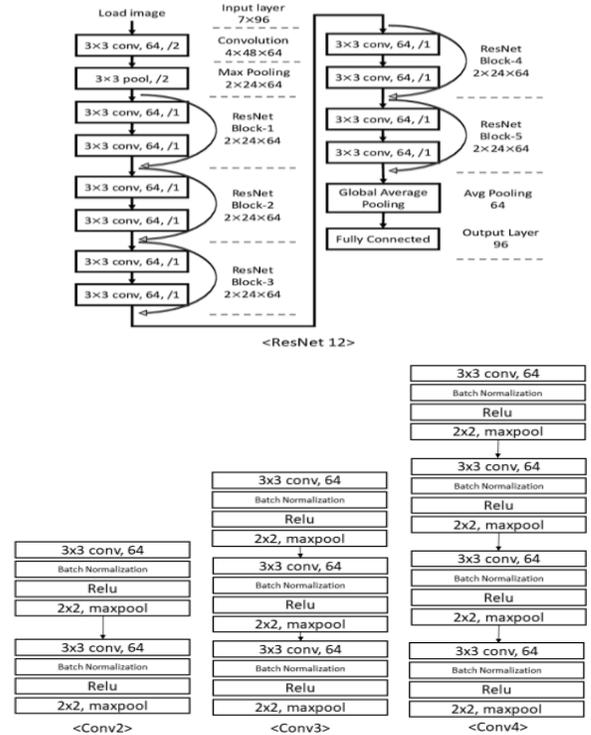

Figure 3. Four Backbone Structures of TablEye

During the preprocessing stage, when encoding categorical features, the use of one-hot encoding results in tabular images composed of only white and black pixels. To make the tabular images appear more natural, we utilized label encoding instead. In Tables 3, 4, and 5, we present the results of our experiments. The abbreviation 'T-A-B' in the 'Method' column signifies a condensed form of 'TablEye-A-B', denoting the implementation of TablEye with 'A' serving as the classifier and 'B' as the backbone. Here, 'P' and 'M' denotes 'Proto-layer' and 'MAML-layer'. 'C2', 'C3', 'C4' and 'R' represents 'Conv2', 'Conv3', 'Conv4' and 'Resnet12'. The term 'Kar' is indicative of the Karhunen dataset, while 'Opt' pertains to the Optdigits dataset.

**Table 3.** Few Shot Tabular Classification Test AUC performance on *3* tabular datasets. We used the AUC performance of XGB, TabNet, SAINT and TabLLM from TabLLM paper. The bold indicates result within *0.01* from highest accuracy. Params indicates the number of parameters for the method. The M-rate signifies the size of the model in comparison to TabLLM. It is calculated as the number of parameters divided by *11* billion.

|        | Diabetes | | | heart | | | car | | | Params | M-rate |
|--------|----------|--------|---------|--------|--------|---------|--------|--------|---------|--------|--------|
| Method | 4-shot | 8-shot | 16-shot | 4-shot | 8-shot | 16-shot | 4-shot | 8-shot | 16-shot | | |
| XGB    | 0.50 | 0.59 | **0.72** | 0.50 | 0.55 | 0.84 | 0.50 | 0.59 | 0.70 | - | - |
| TabNet | 0.56 | 0.56 | 0.64 | 0.56 | 0.70 | 0.73 | ** | 0.54 | 0.64 | - | - |
| SAINT  | 0.46 | 0.65 | **0.73** | 0.80 | **0.83** | **0.88** | 0.56 | 0.64 | 0.76 | - | - |
| TabLLM | 0.61 | 0.63 | 0.69 | 0.76 | **0.83** | 0.87 | **0.83** | **0.85** | **0.86** | **11B** | 1 |
| T-P-R  | 0.68 | **0.70** | 0.69 | 0.72 | 0.78 | 0.69 | 0.69 | 0.68 | 0.75 | 12M | 1/916.67 |
| T-P-C2 | 0.68 | 0.68 | 0.68 | 0.84 | **0.83** | 0.85 | 0.79 | 0.79 | 0.79 | 39K | 1/282051.3 |
| T-P-C3 | 0.71 | 0.73 | 0.71 | **0.86** | 0.79 | 0.78 | 0.72 | 0.71 | 0.76 | 76K | 1/144736.8 |
| T-P-C4 | **0.72** | 0.71 | 0.69 | 0.82 | 0.81 | 0.79 | 0.79 | 0.83 | 0.83 | 113K | 1/97345.13 |

**result omitted due to TabNet package not supporting unseen labels in validation set during cross validation

**Table 2.** Description of Datasets.

| Dataset | Size | # of features | # of classes | Type |
|---------|------|---------------|--------------|------|
| CMC | 1473 | 10 | 3 | N & C |
| Diabetes | 768 | 9 | 2 | N |
| Karhunen | 2000 | 65 | 10 | N |
| Optdigits | 5620 | 65 | 10 | N |
| Car | 1728 | 7 | 4 | C |
| Heart | 918 | 12 | 2 | N |
| Lung | 309 | 16 | 2 | N & C |
| Cancer | 569 | 32 | 2 | N |
| Prostate | 100 | 9 | 2 | N |

## 5.2 Comparison Results with TabLLM

**DATA** The dataset for TabLLM[8] is constrained by token length and the absence of meaningful feature names, which restricts its applicability to datasets such as Karhunen and Optdigits. The datasets utilized in the other experiments, Karhunen and Optdigits, comprised *65* features, rendering TabLLM experiments infeasible. Moreover, these datasets lacked meaningful feature names. Consequently, alternative datasets used in experiments of previous work were selected to replace those. The Diabetes dataset exclusively comprises numerical features, the Heart dataset encompasses both numerical and categorical features, and the Car dataset solely comprises categorical features.

**METRIC** For the Metric, we used the AUC (Area Under the Receiver Operating Characteristic Curve) metric to demonstrate the performance difference compared to the baseline. To compare our method under the same conditions as TabLLM, we used AUC as the metric.

**Shot setting** In the paper of TabLLM, comparisons were made from *4*-shot to *512*-shot. We assume, however, a few-shot scenario, we compared the performance under *4*-shot, *8*-shot, and *16*-shot conditions.

TabLLM transforms tabular data consisting of categorical and numerical features into prompts that can be understood by language model. It leverages the prior knowledge of language models using these prompts. Table 3 displays the performance comparison between our approach and table-to-text method. The experimental results show that for the diabetes dataset composed only of numerical features, TablEye exhibited significantly higher performance in *4*-shot and *8*-shot scenarios compared to table-to-text method and other methods, while XGB[32] and SAINT[33] performed better in the *16*-shot scenario. In *4*-shot, *8*-shot, and *16*-shot scenarios, TabLLM showed performance slightly lower by *0.02* to *0.11* compared to our method. For the heart dataset, our approach outperformed TabLLM by *0.1* in the *4*-shot scenario and showed similar performance in the *8*-shot scenario, but TabLLM outperformed TablEye by *0.02* in the *16*-shot scenario. In the car dataset, TablEye showed slightly lower performance by *0.02* to

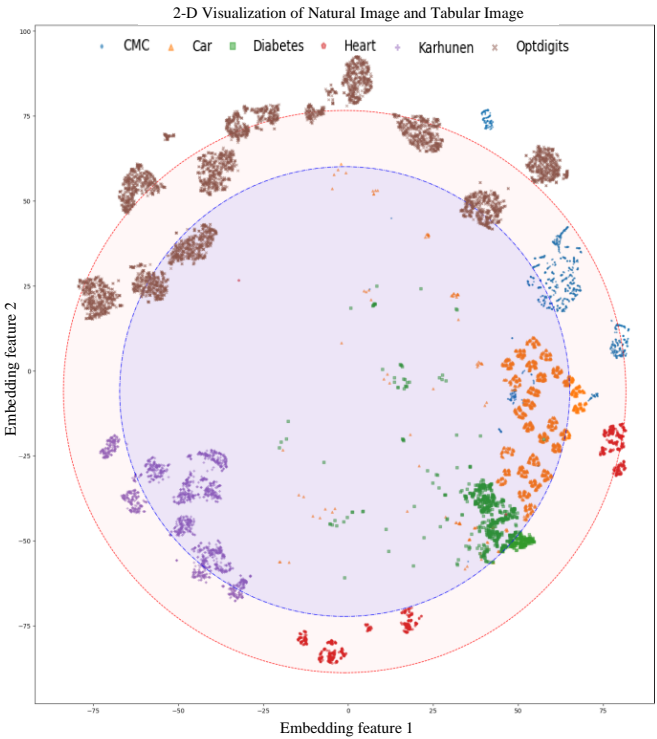

**Figure 4.** 2-D Visualization of Natural Image and Tabular Image Using T-SNE. Each points indicate tabular image, red circle indicates $c_1$ and blue circle indicates $c_2$.

*0.04* compared to TabLLM. Our approach exhibited superior performance to previous work in numeric-only datasets and performed better in *4*-shot scenarios for numeric and categorical datasets but showed similar performance in *8*-shot scenarios. However, in categoric-only datasets, TablEye consistently showed lower performance compared to table-to-text method. We speculate that this is because, due to the nature of previous work utilizing language model, better understands categorical features.

When comparing model sizes, TabLLM has approximately *11 billion* parameters[24]. In contrast, when TablEye uses the heaviest and lowest-performing ResNet12, it has around *12 million* parameters. When using Conv2, Conv3, and Conv4, which showed superior performance, it has a maximum of around *113,000* parameters. A significant difference is observed in the model size, where ResNet12 has parameters approximately *1/916* times the size of those in previous work. Conv2, Conv3, and Conv4 have parameters ranging from *1/97,345* to *1/282,051* times the size. TablEye has a significantly smaller model size compared to the table-to-text method. Our approach also can demonstrate comparable or superior performance and extremely efficient computation power.

## 5.3 Few-Shot Classification Results with Baseline

**Baseline** We chose a supervised learning models that can be experimented within a meta-learning setting without a unlabeled set. For tabular learning, we selected both tree-based model and neural network-based model known for their high performance about tabular learning [34].

**STUNT** A fixed number of unlabeled sets were used as the train set. For the CMC, Diabetes, Karhunen, and Optdigits datasets, *441, 230, 600*, and *1686* unlabeled sets were respectively utilized.

Our experimental analysis was conducted using four public tabular datasets and evaluated the performance of TablEye, which was executed with four backbones and two classifiers. The experiments were performed in both *1*-shot and *5*-shot settings, and the results demonstrate the superiority of TablEye over traditional methods such as XGB and TabNet[35], and even over the STUNT method, which is state of the art about few-shot tabular learning. In the *1*-shot setting, methods of TablEye, T-P-C2 and T-P-C3 exhibited the highest average accuracies of *54.06%* and *54.11%*, respectively, outperforming all other methods. This evidences the robustness of these methods in handling limited data, as they were able to make precise predictions even with minimal training examples.

The performance advantage of TablEye was also evident in the *5*-shot setting, where the T-P-C2 and T-P-C3 methods continued to outperform other methods, achieving average accuracies of *67.72%* and *67.22%*, respectively. This underscores the scalability of these methods, as they showed increased effectiveness with a larger number of labeled examples. Contrastingly, TabNet showed the least performance with an average accuracy of *31.98%* in the *5*-shot setting, indicating its struggle with limited data. Interestingly, XGB, which showed similar struggles in the *1*-shot setting, improved its performance significantly in the *5*-shot setting, achieving an average accuracy of *61.19%*.

**Table 4.** Few Shot Tabular Classification test accuracy(%) on 4 dataset. We report the mean of over 100 iterations. The bold indicates result within *1%* from highest accuracy.

| Method | CMC | Diabetes | Kar | Opt | Avg |
|---|---|---|---|---|---|
| Way | 3 | 2 | 5 | 5 | - |
| Query | 15 | 15 | 15 | 15 | - |
| 1-shot | | | | | |
| XGB | 33.33 | 50.00 | 20.00 | 20.00 | 30.83 |
| TabNet | 34.84 | 51.90 | 21.97 | 20.45 | 32.29 |
| STUNT | 36.52 | 51.60 | 47.72 | 67.92 | 50.94 |
| T-P-R | 35.97 | **58.83** | 30.50 | 44.32 | 42.41 |
| T-P-C2 | **37.33** | 56.53 | **51.21** | **71.18** | **54.06** |
| T-P-C3 | **37.31** | 57.43 | **51.39** | 70.30 | **54.11** |
| T-P-C4 | **37.45** | 57.79 | 44.85 | 65.76 | 51.46 |
| T-M-C2 | 36.60 | **58.34** | 41.92 | 62.04 | 49.73 |
| T-M-C3 | **37.26** | **58.57** | 43.27 | 60.18 | 49.82 |
| T-M-C4 | **37.30** | 57.30 | 43.45 | 60.53 | 49.65 |
| 5-shot | | | | | |
| XGB | **42.18** | 61.20 | 68.21 | 73.19 | 61.19 |
| TabNet | 36.07 | 50.23 | 20.28 | 21.33 | 31.98 |
| STUNT | **41.36** | 55.43 | **83.00** | 86.05 | 66.46 |
| T-P-R | 38.37 | 64.39 | 41.18 | 51.83 | 48.94 |
| T-P-C2 | 40.34 | 65.15 | 77.94 | **87.44** | **67.72** |
| T-P-C3 | **41.22** | 66.20 | 74.61 | **86.83** | **67.22** |
| T-P-C4 | 40.89 | **68.73** | 70.72 | 84.58 | 66.23 |
| T-M-C2 | 37.65 | 63.18 | 56.38 | 62.79 | 55.00 |
| T-M-C3 | 38.48 | 64.35 | 44.80 | 58.79 | 51.60 |
| T-M-C4 | 37.95 | 65.94 | 59.12 | 71.85 | 58.71 |

The STUNT[6] method, which applies a semi-supervised learning approach with a large unlabeled dataset, showed a considerable performance with average accuracies of *50.94%* and *66.46%* in the *1*-shot and *5*-shot settings respectively. The performance of STUNT is, however, heavily influenced by the size and composition of the unlabeled dataset and the configuration of the self-task generation method. In real-world industrial processes, obtaining a sufficiently large and well-composed unlabeled dataset is often challenging, making superior performance of TablEye without relying on unlabeled data highly notable. In summary, our experimental results underscore the effectiveness of TablEye, particularly the T-P-C2 andT-P-C3 methods, in few-shot tabular learning. They consistently outperform traditional machine learning models like XGB and TabNet as well as semi-supervised models like STUNT. Importantly, these advantages are achieved without the need for a large unlabeled dataset, making these methods more practical and applicable in real-world settings where such data might be scarce or hard to obtain.

Our findings also offer insights into the performance of different types of models in few-shot learning scenarios. Specifically, we observed that tree-based models like XGB tend to outperform neural network-based models like TabNet when training data is limited. This suggests that different types of models may be better suited to different learning scenarios, and that the choice of model should be carefully considered based on the specific constraints and requirements of the task at hand. Overall, our work highlights the potential of TablEye and its underlying methods for tackling the challenges of few-shot learning in tabular data classification tasks. By delivering superior performance across a range of settings and datasets, TablEye represents an approach for facilitating more effective and efficient learning from limited data.

### 5.4 Medical Datasets Few-Shot Results

**Data** We obtained medical datasets on diabetes and three types of cancer, which were publicly available on Kaggle. Through these datasets, we aim to validate the applicability of our method to real medical datasets.

**Baseline** Similar to the experiments conducted in the previous baselines, we selected supervised learning models that can be experimented within a meta-learning setting without an unlabeled set. For tabular learning, we chose tree-based model and neural network-based model known for their high performance[34].

**Method** For the method, we chose the approach of utilizing the proto-layer as classifier with conv2, conv3, and conv4 as backbones, which showed overall high performance in the previous experiments using public tabular data.

In the previous experiments, TablEye demonstrated higher accuracy, particularly on numerical datasets, compared to other methods. This paper aims to present the experimental results on four medical datasets to demonstrate the applicability of few-shot tabular learning in the medical domain. Table 5 illustrates the accuracy of the few-shot tabular binary classification test on four medical datasets acquired from Kaggle. Remarkably, T-P-C2, T-P-C3, and T-P-C4 all exhibit superior accuracy compared to both XGB and TabNet, achieving significantly higher mean accuracies in both *1*-shot and *5*-shot scenarios.

In a 1-shot learning scenario, T-P-C4 delivers the highest mean accuracy among all methods at *68.81%*. T-P-C2 and T-P-C3 also present commendable performance, with respective accuracies of *65.98%* and *68.49%*. In a *5*-shot learning context, T-P-C4 maintains its superiority with a mean accuracy of *72.83%*. T-P-C2 attains an accuracy of *71.31%*, surpassing XGB, whereas T-P-C3, with an accuracy of *70.70%*, slightly underperforms compared to XGB. Dissecting the results by individual datasets reveals that while all methods perform well on the Cancer dataset, T-P-C2, T-P-C3, and T-P-C4 outperform the others on the Diabetes and Lung datasets. However, in a *5*-shot scenario, XGB demonstrates the highest accuracy on the Prostate dataset. These results affirm the enhanced performance of our novel few-shot tabular learning framework, represented by T-P-C2, T-P-C3, and T-P-C4, in medical data classification tasks, surpassing conventional methods. The significant increase in mean accuracy for our methods in both *1*-shot and *5*-shot scenarios underscores the efficacy of our approach. Specifically, T-P-C4, exhibiting superior performance when trained with more data (in a *5*-shot setting), further underscores the effectiveness of our method in few-shot learning scenarios.

Table 5. Few Shot Tabular Binary Classification test accuracy(%) on 4 medical datasets from kaggle. We report the mean of over 100 iterations. The bold indicates result within *1%* from highest accuracy.

| Method | Diabetes | Lung | Cancer | Prostate | Avg |
|--------|----------|------|--------|----------|-----|
| Way | 2 | 2 | 2 | 2 | - |
| Query | 15 | 15 | 15 | 15 | - |
| 1-shot | | | | | |
| XGB | 50.00 | 50.00 | 50.00 | 50.00 | 50.00 |
| TabNet | 51.90 | 52.03 | 50.00 | 56.07 | 52.50 |
| T-P-C2 | 56.53 | 63.02 | **85.88** | 58.50 | 65.98 |
| T-P-C3 | **57.43** | 64.93 | 85.08 | **66.51** | 68.49 |
| T-P-C4 | **57.79** | **66.64** | 85.64 | 65.18 | **68.81** |
| 5-shot | | | | | |
| XGB | 61.20 | 63.17 | 81.87 | **78.10** | 71.08 |
| TabNet | 50.23 | 50.43 | 50.17 | 58.57 | 52.35 |
| T-P-C2 | 65.15 | **70.43** | 86.90 | 62.75 | 71.31 |
| T-P-C3 | 66.20 | 65.62 | 87.02 | 63.97 | 70.70 |
| T-P-C4 | **68.73** | 66.69 | **88.92** | 66.98 | **72.83** |

## 6 Discussion

In almost all few-shot scenarios excluding the Car dataset and the Karhunen *5*-shot situation, the TablEye, utilizing Conv2, Conv3, and Conv4 as backbone and a proto-layer as the classifier, demonstrated superior performance. The more complex Resnet12 backbone increased the model size but often underperformed compared to the simpler Conv2, Conv3, and Conv4.

The TablEye displayed impressive performance in *4*-shot comparisons with the Diabetes and Heart datasets, outperforming TabLLM by a significant margin of *0.1* in terms of AUC. Additionally, in *1*-shot scenarios with CMC, Diabetes, Karhunen, and Optdigits datasets, it exhibited an average accuracy *4%* higher than previous studies, and approximately *1%* higher in *5*-shot scenarios. These results indicate that the use of TablEye in few-shot tabular learning problems enables unrestricted experimentation with superior performance compared to traditional research.

TabLLM requires serialization to use tabular data as a prompt. Due to limitations in the token size of the LLM and the necessity for meaningful feature names, it was, however, unable to handle datasets with more than a certain number of features or meaningless feature names, such as the Karhunen and Optdigits datasets[6] with *65* features and feature names like f1, f2, and f3. Therefore, we compared the performance of TablEye and TabLLM using public tabular datasets examined by previous work. The results of our approach confirmed higher performance compared to table-to-text method, particularly in datasets with numerical features such as Diabetes and Heart. Comparing the size of the TabLLM and TablEye previous work possessed a significantly larger number of parameters, requiring considerably higher computational power. Nevertheless, our method demonstrated superior performance with the Diabetes and Heart datasets. Thus, we conclude that our approach is more efficient and showed similar or superior performance for various datasets, overcoming the limitations of TabLLM, which has restrictions on the datasets it can handle and requires high computational power.

STUNT requires a substantial amount of unlabeled data for training. The model used *80*% of the total data as an unlabeled set for training in its paper. In this study, we aimed to use as little unlabeled data as possible to conduct experiments under similar conditions to other baselines, utilizing approximately *30*% of the data as the unlabeled set for training. Despite employing a considerable number of unlabeled sets in experiments with CMC, Diabetes, Karhunen, and Optdigits datasets, TablEye, which did not use any unlabeled sets, showed higher accuracy than STUNT. Increasing the size of the unlabeled set might enhance accuracy of previous work, but obtaining a reasonable unlabeled set is not an easy task. Therefore, we believe that our method has overcome the limitations of STUNT, which requires a large unlabeled set. When applying TablEye to medical datasets, we observed markedly higher accuracy compared to other baselines. For this experiment, we used datasets for diagnosing diabetes and cancer. Compared to existing methods, we achieved an average accuracy of *15*% higher in *1*-shot scenarios and approximately *2*% higher in *5*-shot scenarios. These results indicate that our method can produce meaningful results not only in public tabular data but also in medical data of industrial value.

## 7 Conclusion and Future Work

We proposed TablEye to address the challenges of few-shot tabular learning with prior knowledge learned in image domain. Our approach consisted of two parts: (1) transforming tabular data to image domain. (2) learning prior knowledge from image domain. By transforming data from the tabular domain into the image domain, TablEye generates tabular images, which are then utilized to resolve few-shot tabular learning problems. The backbone and classifier are trained on natural image. They predict result about query tabular images using support tabular images.

Performance of TablEye was compared with several other baseline methods, such as TabLLM and STUNT. In these comparisons, our method consistently displayed superior performance, outpacing the other methods on both public tabular data and medical data. Notably, our approach effectively overcomes several limitations inherent to traditional few-shot tabular learning methods. These limitations include a dependence on the number and names of features in the dataset, the need for substantial computational power, and the requirement for a large unlabeled set.

Although TablEye demonstrates exceptional performance with numerical features, it exhibits relatively lower performance with categorical features. We posit that this disparity is due to our domain transformation method not being optimized for categorical values. Future work will explore the incorporation of few-shot tabular learning methods based on Large Language Model (LLM), which show particular efficacy with categorical features. This method will be integrated using ensemble or multimodal approaches, thereby allowing each method to handle the type of feature it is most adept at processing. The innovative approach taken by TablEye provides a solid foundation for future research. We believe that leveraging the image domain to solve problems in the tabular domain opens exciting new possibilities for advancing the field of tabular learning. As such, we anticipate that future research will continue to explore and expand upon the ideas and techniques introduced by TablEye.

## Appendix. A. Pseudo Code

```python
import numpy as np
# "lower_triangular_indices " gets indices for the lower-triangle of the matrix.
def pixel_distance_ranking(num_r, num_c): # Generate a pixel distance matrix Q for domain transformation
    coord_rows, coord_cols = np.meshgrid(np.arange(num_r), np.arange(num_c), indexing='ij')
    coordinate = np.vstack((coord_rows.ravel(), coord_cols.ravel())).T
    # Calculate the Euclidean distance."diff"'s  each plane contains the difference between a pair of coordinates.
    diff = coordinate[:, np.newaxis, :] - coordinate[np.newaxis, :, :]
    cord_dist = np.sqrt(np.sum(diff**2, axis=-1))
    # Generate a ranking based on the distance.
    tril_id = lower_triangular_indices (num_r * num_c)
    rank = rankdata(cord_dist[tril_id])
    # Build the final distance ranking matrix. Populate the lower triangle of the ranking matrix using the computed ranks, and then add its transpose to itself to get the full symmetric matrix.
    ranking = np.zeros_like(cord_dist)
    ranking[tril_id] = rank
    ranking += ranking.T
    return coordinate, ranking

def feature_distance_ranking(data):# Generate a feature distance matrix R for domain transformation
    #'calculate_euclidean_distance' computes the distance between each pair of columns in 'data'.
    corr = calculate_euclidean_distance(data)
    # 'dissimilarity' is 1 minus the distance, which effectively measures the similarity between features. The closer the value is to 1, the more similar the features are.
    dissimilarity = round(1 - corr, decimals=10)
    # Generate a ranking based on the dissimilarity. This ranking treats smaller dissimilarity as higher ranks.
    tril_id = lower_triangular_indices(data.shape[1])
    rank = rankdata(dissimilarity[tril_id])
    # Build the final dissimilarity ranking matrix. Populate the lower triangle of the ranking matrix using the computed ranks, and then add its transpose to itself.
    ranking = zeros_like(dissimilarity)
    ranking[tril_id] = rank
    ranking += ranking.T
    return ranking, corr
```

## Appendix. B. Reproducibility

To reproduce the framework proposed in this paper, the few-shot learning task process was implemented using the LibFewShot library [36]. A configuration file for utilizing the LibFewShot library and detailed model setting for reproduction will be made publicly available on GitHub.

# REFERENCES

[1] Wang, Y., Yao, Q., Kwok, J. T., & Ni, L. M. (2020). Generalizing from a few examples: A survey on few-shot learning. *ACM computing surveys (csur)*, *53*(3), 1-34.

[2] Chen, Z., Maji, S., & Learned-Miller, E. (2021). Shot in the dark: Few-shot learning with no base-class labels. In *Proceedings of the IEEE/CVF Conference on Computer Vision and Pattern Recognition,* 2668-2677.

[3] Min, S., Lewis, M., Hajishirzi, H., & Zettlemoyer, L. (2021). Noisy channel language model prompting for few-shot text classification. *arXiv preprint arXiv:2108.04106*.

[4] Deng, J., Dong, W., Socher, R., Li, L. J., Li, K., & Fei-Fei, L. (2009, June). Imagenet: A large-scale hierarchical image database. In *2009 IEEE conference on computer vision and pattern recognition*, 248-255.

[5] Mathov, Y., Levy, E., Katzir, Z., Shabtai, A., & Elovici, Y. (2020). Not all datasets are born equal: On heterogeneous data and adversarial examples. *arXiv preprint arXiv:2010.03180*.

[6] Nam, J., Tack, J., Lee, K., Lee, H., & Shin, J. (2023). STUNT: Few-shot Tabular Learning with Self-generated Tasks from Unlabeled Tables. *arXiv preprint arXiv:2303.00918*.

[7] Van Engelen, J. E., & Hoos, H. H. (2020). A survey on semi-supervised learning. *Machine learning*, *109*(2), 373-440.

[8] Hegselmann, S., Buendia, A., Lang, H., Agrawal, M., Jiang, X., & Sontag, D. (2023, April). Tabllm: Few-shot classification of tabular data with large language models. In *International Conference on Artificial Intelligence and Statistics*, 5549-5581.

[9] Zhou, Z. H. (2018). A brief introduction to weakly supervised learning. *National science review*, *5*(1), 44-53.

[10] Pan, S. J., & Yang, Q. (2010). A survey on transfer learning. *IEEE Transactions on knowledge and data engineering*, *22*(10), 1345-1359.

[11] Liu, B., Wang, X., Dixit, M., Kwitt, R., & Vasconcelos, N. (2018). Feature space transfer for data augmentation. In *Proceedings of the IEEE conference on computer vision and pattern recognition*, 9090-9098.

[12] Zhao, L., Pan, S., Xiang, E., Zhong, E., Lu, Z., & Yang, Q. (2013, June). Active transfer learning for cross-system recommendation. In *Proceedings of the AAAI Conference on Artificial Intelligence*,27(1), 1205-1211.

[13] Li, P., Cui, H., Khan, A., Raza, U., Piechocki, R., Doufexi, A., & Farnham, T. (2021, May). Deep transfer learning for WiFi localization. In *2021 IEEE Radar Conference (RadarConf21),* 1-5.

[14] Nathani, N., & Singh, A. (2021). Foundations of Machine Learning. In *Introduction to AI Techniques for Renewable Energy Systems*, 43-64.

[15] Vapnik, V. (1991). Principles of risk minimization for learning theory. *Advances in neural information processing systems*, *4*.

[16] Bottou, L., & Bousquet, O. (2007). The tradeoffs of large scale learning. *Advances in neural information processing systems*, *0*.

[17] Bottou, L., Curtis, F. E., & Nocedal, J. (2018). Optimization methods for large-scale machine learning. *SIAM review*, *60*(2), 223-311.

[18] Nguyen, H., & Zakynthinou, L. (2018). Improved algorithms for collaborative PAC learning. *Advances in Neural Information Processing Systems*, *31*.

[19] Mahadevan, S., & Tadepalli, P. (1994). Quantifying prior determination knowledge using the pac learning model. *Machine Learning*, *17*, 69-105.

[20] Germain, P., Bach, F., Lacoste, A., & Lacoste-Julien, S. (2016). PAC-Bayesian theory meets Bayesian inference. *Advances in Neural Information Processing Systems*, *29*.

[21] Snell, J., Swersky, K., & Zemel, R. (2017). Prototypical networks for few-shot learning. *Advances in neural information processing systems*, *30*.

[22] Brown, T., Mann, B., Ryder, N., Subbiah, M., Kaplan, J. D., Dhariwal, P., ... & Amodei, D. (2020). Language models are few-shot learners. *Advances in neural information processing systems*, *33*, 1877-1901.

[23] Liu, H., Tam, D., Muqeeth, M., Mohta, J., Huang, T., Bansal, M., & Raffel, C. A. (2022). Few-shot parameter-efficient fine-tuning is better and cheaper than in-context learning. *Advances in Neural Information Processing Systems*, *35*, 1950-1965.

[24] Sanh, V., Webson, A., Raffel, C., Bach, S. H., Sutawika, L., Alyafeai, Z., ... & Rush, A. M. (2021). Multitask prompted training enables zero-shot task generalization. *arXiv preprint arXiv:2110.08207*.

[25] Finn, C., Abbeel, P., & Levine, S. (2017, July). Model-agnostic meta-learning for fast adaptation of deep networks. In *International conference on machine learning*, 1126-1135.

[26] Zhang, Z., & Sabuncu, M. (2018). Generalized cross entropy loss for training deep neural networks with noisy labels. *Advances in neural information processing systems*, *31*.

[27] Vinyals, O., Blundell, C., Lillicrap, T., & Wierstra, D. (2016). Matching networks for one shot learning. *Advances in neural information processing systems*, *29*.

[28] Zhu, Y., Brettin, T., Xia, F., Partin, A., Shukla, M., Yoo, H., & Stevens, R. L. (2021). Converting tabular data into images for deep learning with convolutional neural networks. *Scientific reports*, *11*(1), 11325.

[29] Vanschoren, J., Van Rijn, J. N., Bischl, B., & Torgo, L. (2014). OpenML: networked science in machine learning. *ACM SIGKDD Explorations Newsletter*, *15*(2), 49-60.

[30] He, K., Zhang, X., Ren, S., & Sun, J. (2016). Deep residual learning for image recognition. In *Proceedings of the IEEE conference on computer vision and pattern recognition*, 770-778.

[31] Van der Maaten, L., & Hinton, G. (2008). Visualizing data using t-SNE.*Journal of machine learning research*, *9*(11).

[32] Chen, T., & Guestrin, C. (2016, August). Xgboost: A scalable tree boosting system. In *Proceedings of the 22nd acm sigkdd international conference on knowledge discovery and data mining*, 785-794.

[33] Somepalli, G., Goldblum, M., Schwarzschild, A., Bruss, C. B., & Goldstein, T. (2021). Saint: Improved neural networks for tabular data via row attention and contrastive pre-training. *arXiv preprint arXiv:2106.01342*.

[34] Shwartz-Ziv, R., & Armon, A. (2022). Tabular data: Deep learning is not all you need. *Information Fusion*, *81*, 84-90.

[35] Arik, S. Ö., & Pfister, T. (2021, May). Tabnet: Attentive interpretable tabular learning. In *Proceedings of the AAAI Conference on Artificial Intelligence*, 35(8), 6679-6687.

[35] Arik, S. Ö., & Pfister, T. (2021, May). Tabnet: Attentive interpretable tabular learning. In *Proceedings of the AAAI Conference on Artificial Intelligence*, 35(8), 6679-6687.

[36] Li, W., Yang, X., Dong, C., Tian, P., Qin, T., Huo, J., ... & Luo, J. (2021). Libfewshot: A comprehensive library for few-shot learning. *arXiv preprint arXiv:2109.04898*.